\documentclass[10pt,twocolumn,letterpaper]{article}

\usepackage[pagenumbers]{iccv} 
\usepackage{multirow} 
\usepackage{xcolor}
\usepackage{pifont}
\usepackage{amsmath}
\usepackage{colortbl}
\usepackage{makecell}
\definecolor{bestcolor}{HTML}{F8CBAD} 
\definecolor{bestcolormetric}{HTML}{8684FF} 
\usepackage{colortbl}
\usepackage{afterpage} 

\definecolor{iccvblue}{rgb}{0.21,0.49,0.74}
\usepackage[pagebackref,breaklinks,colorlinks,allcolors=iccvblue]{hyperref}

\title{You Think, You ACT: The New Task of Arbitrary Text to Motion Generation}

\author{Runqi Wang$^{1,2 *}$\quad 
Caoyuan Ma$^{1,2 *}$\quad 
Guopeng Li$^{3 *}$\quad 
Hanrui Xu$^2$\quad 
Yuke Li$^4$\quad 
Zheng Wang$^{1,2 \dag}$\quad
\\
$^1$National Engineering Research Center for Multimedia Software, Wuhan University\\
${^2}$School of Computer Science, Wuhan University \quad ${^3}$StepFun\\
${^4}$University of Maryland College Park\\
{\small $*$ equal contribution, $\dag$ corresponding to wangzwhu@whu.edu.cn} 
}

\begin{document}
\twocolumn[{%
\renewcommand\twocolumn[1][]{#1}%
\vspace{-8mm}
\maketitle
\begin{center}
    \centering
    \vspace{-8mm}
    \includegraphics[width=\textwidth]{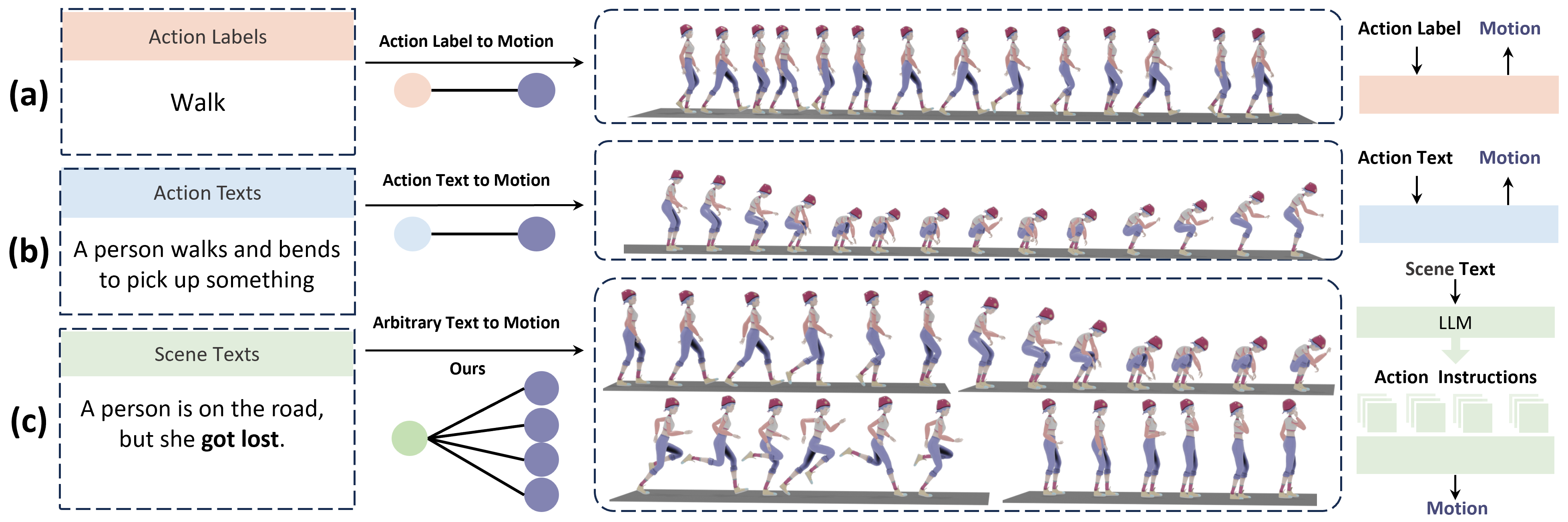}
     \vspace{-4mm}
      \captionof{figure}{
  (a) ``Action Label to Motion'' generates motion with a certain action label.
  (b) ``Action Text to Motion'' generates motion with an explicit Action Text (texts contain action labels). 
  (c)  ``Arbitrary Texts to Motion'' encompasses a new Scene Text to Motion task, where scene text refers to events or situations and does not contain explicit action labels. Understanding these scene texts and generating corresponding reactive motions is a multi-solution task.
      }
       \label{fig:teaser}
\end{center}%
}]
\begin{abstract}
Text to Motion aims to generate human motions from texts. 
 Existing settings rely on limited Action Texts that include action labels (\emph{e.g.}, ``walk, bend''), which limits flexibility and practicability in scenarios difficult to describe directly.
This paper extends limited Action Texts to arbitrary ones. Scene texts without explicit action labels can enhance the practicality of models in complex and diverse industries such as virtual human interaction, robot behavior generation, and film production, while also supporting the exploration of potential implicit behavior patterns.
However, newly introduced Scene Texts may yield multiple reasonable output results, causing significant challenges in existing data, framework, and evaluation.

To address this practical issue, we first create a new dataset \textsc{HumanML3D++} by extending texts of the well-annotated dataset \textsc{HumanML3D}.
Secondly, we propose a simple yet effective framework that extracts action instructions from arbitrary texts and subsequently generates motions. 
Furthermore, we also benchmark this new setting with multi-solution metrics to address the inadequacies of existing single-solution metrics. 
Extensive experiments indicate that Text to Motion in this realistic setting is challenging, fostering new research in this practical direction. 
Our data, model, and code will be released.
\end{abstract}    
\section{Introduction}
\label{sec:intro}

Text to Motion (T2M) denotes generating motions from natural language,
reducing labor costs in industries requiring motion capture and manual editing.
While current T2M methods have demonstrated effectiveness in controllable settings, 
some real-world applications such as Open-world games and virtual assistants often demand greater flexibility, where characters must dynamically respond to unrestricted and diverse scene inputs from users, rather than following a direct action command.
For example, in an open-world game, a player might warn an NPC, ``Watch out! Left punch incoming!'' The NPC must interpret the intent and dodge or block, rather than execute a `` punch'' command.
Despite this need, existing T2M methods still relies on explicit action information (such as Action Labels~\cite{ACTOR,guo2020action2motion,ODMO,yu2020structure,lucas2022posegpt,degardin2022generative} and Action Texts~\cite{ahuja2019language2pose,zhang2022motiondiffuse,MDM,t2m-gpt,guo2024momask,tevet2022motionclip,temos} in Figure~\ref{fig:teaser}(a-b)), which are difficult to apply in less directly describable and more flexible settings, as illustrated in Figure~\ref{fig:teaser}(c). 
Therefore, exploring the generation of potential motions from arbitrary texts is important.

Generating motion from arbitrary texts (Arbitrary Text to Motion) includes scene texts that describe events or situations without explicit action labels, contrasting with the ``Action Text (label) to Motion'' approach~\cite{mld,MDM,zhang2022motiondiffuse,kong2023priority,jin2024act}, which is a fixed one-to-one alignment of a precise action text and a certain kind of motion pattern (e.g., a description ``walking'' directly corresponds to walking pattern). Instead, ``Arbitrary Text to Motion'' introduces a multi-solution paradigm, allowing multiple plausible motion patterns to be generated from a single scene text (e.g., in Figure~\ref{fig:teaser}(c), a scene text can generate four (``walk'', ``run'', ``bend down'', ``take a photo'') different motion patterns). Consequently, datasets, frameworks, and evaluation metrics designed for deterministic one-to-one alignment in ``Action Text to Motion'' are unsuitable for new \textbf{multi-solution} task.

\textbf{Dataset.}
As shown in Table~\ref{tab:datasetstatistics}, existing datasets~\cite{mahmood2019amass,ji2018large,shahroudy2016ntu,punnakkal2021babel,guo2020action2motion,Plappert2016} primarily focus on explicit Action Text annotations to describe motions, while Scene Text remains scarce.
Furthermore, the multi-solution characteristics of Scene Text pose challenges in semantic correspondence.
To address these issues, we introduce the~\textsc{HumanML3D++} by adding Scene Texts to evaluate the ``Arbitrary Text to Motion".
To ensure both the scale and semantic coherence with motions, we use a LLM with a carefully designed prompting strategy and causal contextual guidance to generate Scene Texts.
For data quality, we employ a hybrid denoising process that integrates automated filtering with manual verification. 
Our dataset comprises about 45k Action Texts, 135k Scene Texts, and 15k motion sequences, providing a foundation for the ``Arbitrary Text to Motion".

\begin{table}[t]
\centering
\renewcommand{\arraystretch}{1.5}
\resizebox{0.44\textwidth}{!}{%
\begin{tabular}{cclccccc}
\toprule
\multicolumn{1}{c}{}&\multicolumn{1}{c}{}& \multicolumn{1}{l}{\textbf{ Name}} & \multicolumn{1}{c}{\textbf{Sub.}} & \multicolumn{1}{c}{\textbf{Motion}} & \multicolumn{1}{c}
{\textbf{Text}} & \multicolumn{1}{c}{\textbf{Act. Class}} & \multicolumn{1}{c}{\textbf{Scene}} \\ 
\midrule
\multirow{9}{*}{\rotatebox{90}{\quad  \quad \quad \quad Action Label}}
&\multirow{9}{*}{\rotatebox{90}{\quad \quad \quad \quad  Dataset}}
& AMASS~\cite{mahmood2019amass} &344  & 11k &  $-$   & $-$ & \ding{55}\\
& & NTU-120RGB+D\cite{liu2019ntu} &106 & 114k& $-$  & 120 & \ding{55} \\
& &UESTC~\cite{ji2018large} & 118 & 26k& $-$&40 & \ding{55} \\ 
& & NTU RGB+D~\cite{shahroudy2016ntu} & $-$  &  56k  &$-$  & 60 &\ding{55} \\
& & BABEL~\cite{punnakkal2021babel}& 344 &66k & $-$ & 250 & \ding{55}\\
& & HumanAct12~\cite{guo2020action2motion}& 12 &1k & $-$  & 12 &\ding{55}\\
\midrule
\multirow{3}{*}{\rotatebox{90}{Text }}& \multirow{3}{*}{\rotatebox{90}{Dataset}}& 
KIT-ML~\cite{Plappert2016} & 111 &3k & 6k & $-$ & \ding{55} \\
& & HumanML3D~\cite{guo2022generating} & 344  & 15k  & 45k & $-$  & \ding{55} \\
& & \cellcolor{bestcolor}{Ours} & \cellcolor{bestcolor}{344} &\cellcolor{bestcolor}{15k} & \cellcolor{bestcolor}{\textbf{135k}}& \cellcolor{bestcolor}{$-$} & \cellcolor{bestcolor}{\textcolor{red}{\ding{51}}}\\
\bottomrule
\end{tabular}
}%
\vspace{-2mm}
\caption{\textbf{Dataset comparison.} \textit{Sub.} refers to the number of humans included and \textit{Act. Class} denotes the variety of action classes (not applicable to datasets with Action Texts). Our dataset excels with the most extensive annotated text content, especially with a substantial amount of Scene Texts.}
\vspace{-4mm}
 \label{tab:datasetstatistics}
\end{table}

\textbf{Framework.}
Previous coupled frameworks~\cite{temos,mld, MDM,guo2024momask,t2m-gpt,zhang2022motiondiffuse,zhong2023attt2m} are limited to a simple and deterministic mapping between Action Text and motion, struggling to handle the complex Scene Text semantic parsing and motion plausibility.
We propose a~\textbf{T}hink and \textbf{A}ct framework for \textbf{A}rbitrary \textbf{T}ext (TAAT).
During the Think Stage, we leverage the optimal solution for multi-solution problems, \emph{i.e.}, LLM for cognitive processing and scene interpretation, exploring multiple action instructions derived from a single textual input.
In the Act Stage, we improve the transformer to enable deterministic motion generation, avoiding the omission of action.
Consequently, our framework decouples both cognitive processing and action execution, breaking the limitations of traditional T2M paradigms.

\textbf{Evaluation.}
The existing evaluation metrics fail to accommodate multi-solution scenarios.
Specifically, metrics(\emph{e.g.} ``R-Precision, MM-Dist''~\cite{guo2022generating}) assume a unique ground truth for generated motions, neglecting the possibility of output motions that match the Scene Text but differ from the dataset’s GT.
We further validate the dataset's ground truth with existing metrics to demonstrate that ``R-Precision, MM-Dist'' cannot effectively evaluate multi-solution issues (Sec~\ref{sec:metric}).
We introduce two novel metrics, Hit Accuracy and Mean Hit Distance, which are designed to assess the validity of generated results in a manner aligned with the multi-solution characteristic of our approach, along with a comprehensive model performance analysis of different models across various application settings.

Our main contributions can be summarized as follows:
\begin{itemize}
    \item We first introduce the Arbitrary Text to Motion task, expanding existing Text to Motion to more practical and flexible settings.
    \item We construct a comprehensive dataset with over 135k Scene Text annotations and propose a novel think-and-act framework to infer potential motions from Scene Texts.
   \item We develop a multi-solution evaluation metric system. Extensive experiments demonstrate our method generates more coherent motions than existing approaches.
\end{itemize}
\begin{figure*}[ht!]
    \centering
    \includegraphics[width=\textwidth]{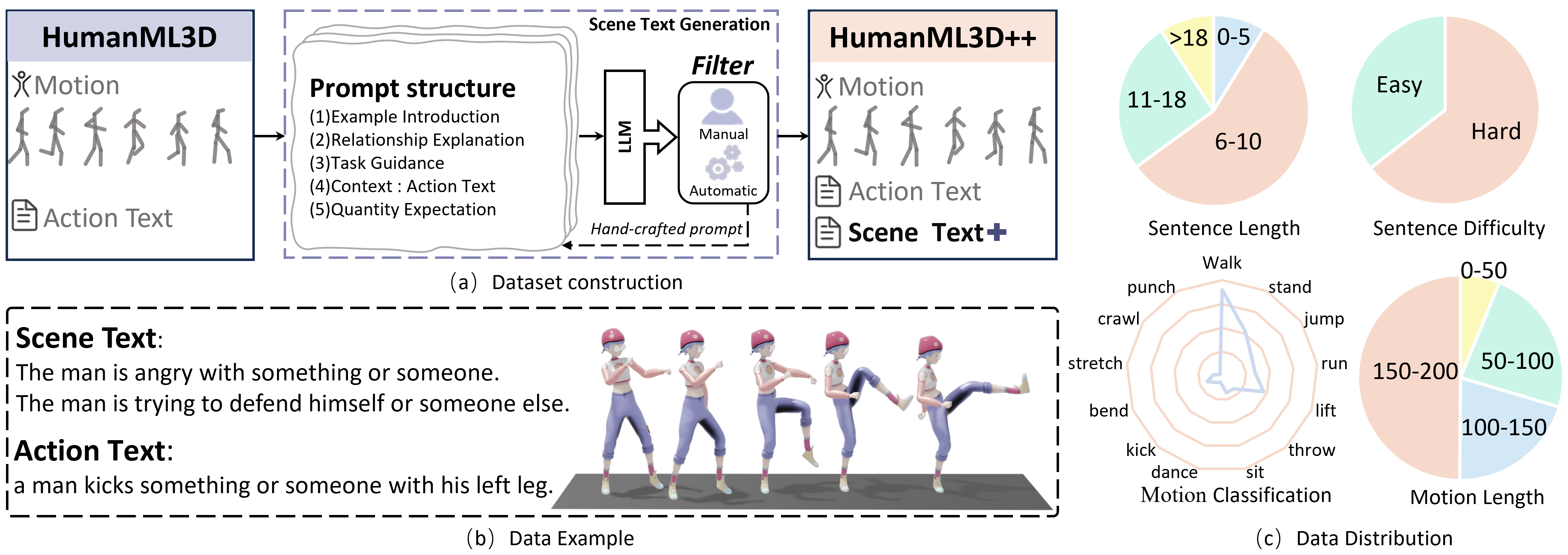}
      \vspace{-8mm}
    \caption{\textbf{Dataset overview.} (a) We construct a novel \textsc{HumanML3D++} dataset through causal contextual guiding prompts and post-filtering. 
    This is the first dataset with dual-text support.
    (b) Detailed data examples.
    }
    \label{fig:dataset_total}
    \vspace{-4mm}
\end{figure*}

\section{Related Work}
\label{sec:Related Work}

Human motion generation supports diverse multimodal inputs, including Action Texts~\cite{ahuja2019language2pose,zhang2022motiondiffuse,MDM,t2m-gpt,guo2024momask,tevet2022motionclip,temos,zhou2024avatargpt}, action labels~\cite{ACTOR,guo2020action2motion,ODMO,yu2020structure,lucas2022posegpt,degardin2022generative},incomplete posture~\cite{duan2021single,harvey2020robust,MDM},music~\cite{lee2019dancing,li2022danceformer,li2021aidancegeneration,aristidou2022rhythm,sun2022you}, images~\cite{rempe2021humor}, and more~\cite{wang2024move,jiang2024scaling,tanaka2023role,liang2024intergen,wang2023intercontrol}.
In common human motion generation, deterministic, action-centric texts serve as the primary inputs. 
Our task shifts to scene-based texts, enabling more responsive and adaptable motions from general descriptions. 
This novel perspective broadens the scope of T2M, increasing flexibility and enhancing usability.

\subsection{Text to Motion Generation}

Arbitrary Text to Motion broadens traditional Text to Motion task, accommodating more diverse and flexible scenarios.  
Previous Text to Motion methodologies typically establish a deterministic one-to-one mapping between text and motion.
Existing methods can be grouped into two main categories. Diffusion-based methods~\cite{mld,MDM,zhang2022motiondiffuse,kong2023priority,jin2024act,wang2023fg,karunratanakul2023guided} utilize long Markov chains of reverse diffusion steps to generate samples, which can be computationally expensive and slow. Additionally, they cannot produce sequences of variable lengths. 
Furthermore, fine-grained motion generation methods~\cite{wang2023fg,karunratanakul2023guided} require lengthy and complex training data or additional inputs like obstacles, trajectories, and keyframes, compromising practicality.
Other approaches~\cite{guo2022tm2t,t2m-gpt,lu2023humantomato,zhong2023attt2m,guo2024momask}use VAEs and transformers to model sequence relationships. 
However, due to limitations in model scale and training data, transformers may demonstrate suboptimal performance on unseen samples, resulting in unclear outputs.
Additionally, HUMANTOMATO~\cite{lu2023humantomato} requires pre-training text-motion alignment models on specific datasets, while Momask~\cite{guo2024momask} presents motions that are constrained and lack diversity, necessitating a target length input for generation and a pre-trained text-to-length model for sampling.
Our approach leverages transformers' temporal modeling capabilities and LLMs' robust language generation and zero-shot transfer abilities. 
This synergy enables effective handling of complex textual inputs without the need for additional training data or auxiliary models.

\subsection{Extended Text to Motion Generation}
Recently, some research has incorporated interactive factors into the tasks of human motion generation. For instance, ~\cite{wang2024move,jiang2024scaling,li2024genzi,xu2023interdiff,Pi_2023_ICCV} integrated environmental elements into human motion generation, while ~\cite{tanaka2023role,liang2024intergen,wang2023intercontrol,xu2024inter,diller2023cghoi,ghosh2024remos} utilized textual guidance to facilitate the generation of interactive motions for pairs or groups. 
However, 3D data is challenging to obtain and lacks user-friendliness. Moreover, certain emotional or event states cannot be directly represented through 3D data.
Compared to other methods, our approach offers greater flexibility in inputs and can cover a wider range of scenarios and contexts.

\begin{figure*}[h!]
    \centering
    \includegraphics[width=\textwidth]{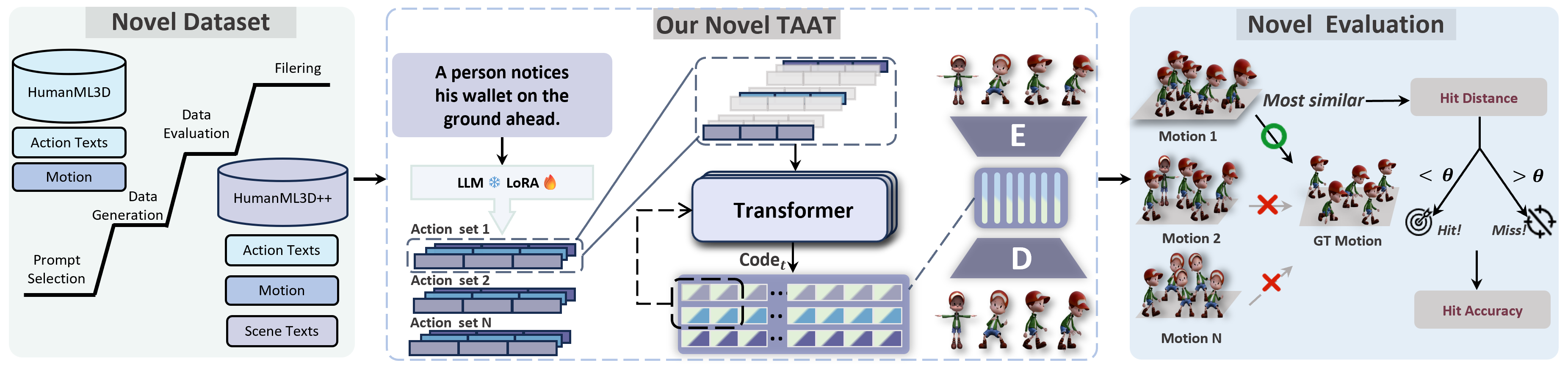}
      \vspace{-7mm}
    \caption{\textbf{Pipeline overview.} 
 (Dataset) We extend \textsc{HumanML3D}~\cite{guo2022generating} to the novel \textsc{HumanML3D++} with Scene Texts.
 (TAAT) We utilize a fine-tuned LLM to generate multiple reasonable response action instructions for a single Scene Text.
 And we generate each action in an action instruction individually and utilize the code generated in the previous stage to guide the generation in the subsequent stage.
(Evaluation) We introduce two new metrics, Hit Accuracy and Mean Hit Distance, to measure this multi-solution task.
    }
    \label{fig:mehtod}
    \vspace{-4mm}
\end{figure*}

\section{Dataset: \textsc{HumanML3D++} }
\label{sec:dataset}
We introduce \textsc{HumanML3D++}, a novel dataset designed to support more flexible Arbitrary Text to Motion by providing both Action and Scene Texts for motion data. 
As shown in Table~\ref{tab:datasetstatistics}, \textsc{HumanML3D++} includes over 135k scene texts paired with matching motions, making it the first annotated motion dataset with dual-text support.

\textbf{Dataset Construction.}
(1) Data Collection: we select HumanML3D~\cite{guo2022generating} as our foundation dataset.
The selection is based on the dataset's inclusion of high-quality action annotations and a large volume of motion data, aiming to reduce annotation costs and ensure diversity.
(2) Prompt Strategy Design: we design different prompts and employ expert scoring to assess the generated outcomes of various prompts.
A comparison under different prompts is presented in Table~\ref{tab:prompt}. Ultimately, we choose the causal context-guided prompt as the final strategy for generating data.
(3) Data Generation: For each motion in the source dataset,  which comprises 3-5 Action Texts, we use LLMs to generate more than six Scene Texts per motion to ensure comprehensive data coverage due to the inherent multi-solution nature of the problem.
\textbf{Data Quality Control.} 
To ensure high quality and reliability of the data, we implement a series of denoising and validation measures.
(1) Filtering of Invalid Content: We use length checks and fixed-format matching to filter out invalid generated content.
(2) Text Formatting: All texts are processed uniformly according to the format standards of \textsc{HumanML3D}.
(3) Cross-validation with Large Language Models: We employ a LLM, independent from the data collection phase, for cross-validation. This LLM comprehensively evaluates the generated scene text annotations based on predefined criteria, determining their appropriateness and flagging any anomalous texts for review.
(4) Manual Evaluation and Iterative Optimization: We invite 20 participants to assess a randomly selected subset of data (15\% of the total). All participants undergo standardized training. Each scene text is evaluated by at least three participants. A scene text is deemed acceptable if all three evaluators agree it is contextually appropriate, free from ambiguity and mismatches. The iterative improvement process continues until the validation accuracy exceeds 95\%.
\begin{figure}[t!]
    \centering
    \includegraphics[width=0.46\textwidth]{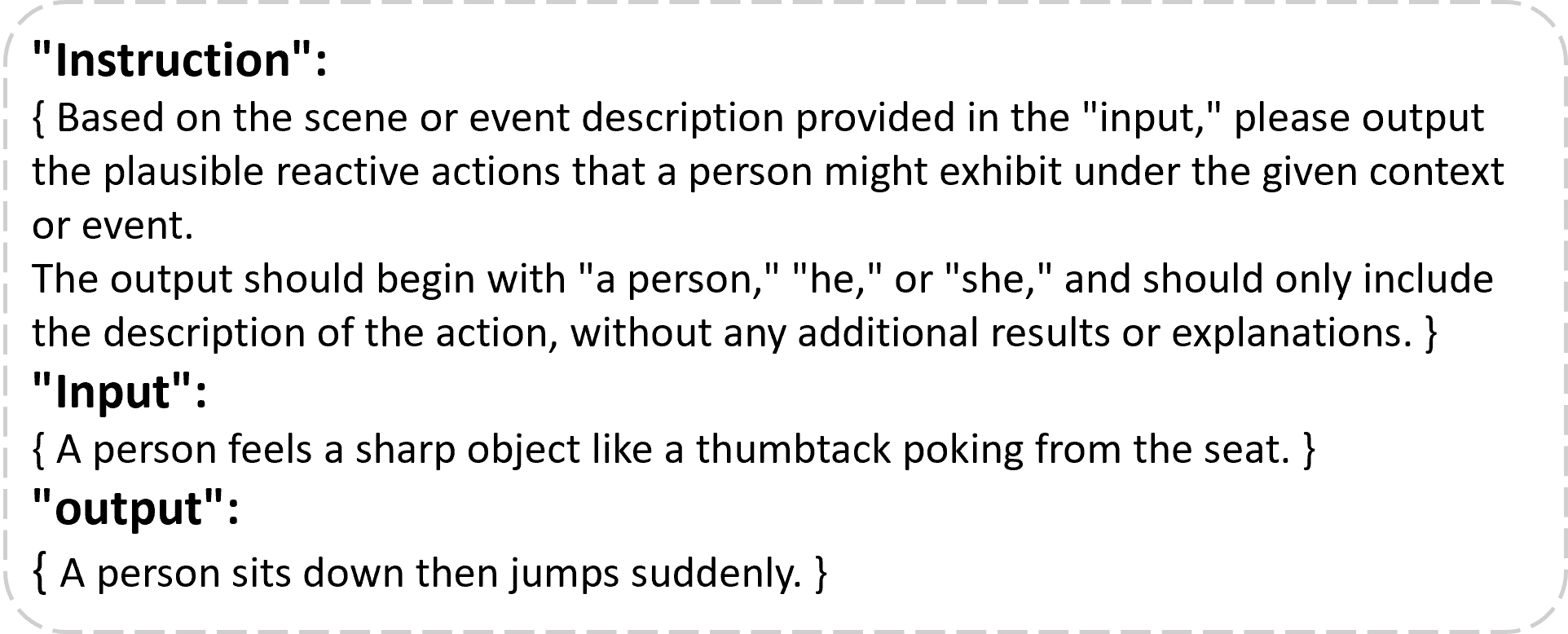}
    \vspace{-3mm}
    \caption{Examples of the fine-tuning data we used.}
    \vspace{-4mm}
    \label{fig:fintune_case1}
\end{figure}

\section{Method}
Given scene or action textual inputs, our objective is to generate realistic human motion \(\mathbf{X} = [x_1, x_2, \ldots, x_t] \), where each \( x_t \in \mathbb{R}^d \) represents the human body pose in a \( d \)-dimensional space at frame \( t \).
As illustrated in Figure~\ref{fig:mehtod}, the framework adopts a dual-phase architectures. 
In the Think phase, we harness the emergent properties of large language models to delineate the underlying relationship between textual inputs and action instructions.
We synthesize the temporal sequence modeling capabilities inherent in Transformers with discrete action representations, establishing a bijective relationship in the Act phase.
This dual-phase architecture decouples text comprehension from motion generation, enabling thorough scene understanding before generating precise motion sequences, 
which is effective for handling the multi-solution nature of the task.

\subsection{Think Model}
Modeling and learning from arbitrary texts, especially Scene Texts, differ significantly from previous tasks due to the multi-solution nature.
This can be mathematically formulated as  \( p(\mathbf{X} \mid \mathbf{T}) \), where \(\mathbf{T}\) represents the arbitrary text and \(\mathbf{X}\) represents the corresponding motion sequence.
The mapping from \(\mathbf{T}\) to \(\mathbf{X}\) is conditional and non-deterministic, allowing multiple motion sequences to arise from the same text.
For this complex situational comprehension, large language models represent an optimal approach due to their advanced linguistic understanding and heightened contextual sensitivity when engaging with complex textual data.
Through the refinement of pre-trained architectures using structured input-output pairs that encompass Scene Texts and associated responses, we can more adeptly leverage the vast training data inherent to LLMs.
Formally, the objective can be defined as minimizing the following loss function over the dataset \(\mathcal{D}\):
\begin{small}
\begin{equation}
\mathcal{L}_{\text{LLM}} = \sum_{(Q, A) \in \mathcal{D}} \mathbb{E}_Q \left[ \log p(A \mid \theta_{\text{LLM}}(Q) \right],
\end{equation}
\begin{equation}
\mathcal{D} = \{(Q_i, A_i)\}_{i=1}^{N},
\end{equation}
\end{small}
where \(p(A \mid Q; \theta_{\text{LLM}})\) represents the probability of generating the correct response \(A\) given the question \(Q\) under the LLM parameters \(\theta_{\text{LLM}}\).
\(\mathcal{D}\) denotes the data used for fine-tuning, with \(Q\) is Scene Text input and \(A\) is Action Texts GT output. 
To mitigate resource consumption, we employ LoRA~\cite{hu2022lora} to fine-tune the LLM. 
In alignment with established practices for large language models, we utilize cross-entropy loss to enforce similarity between the predicted tokens $\mathcal{A}$ and the ground truth tokens $\mathcal{A}^{gt}$, formally represented as follows:
\begin{small}
\begin{equation}
\mathcal{L}_{\text{token}} = \mathrm{CE}(\mathcal{A}, \mathcal{A}^{\text{gt}}) = - \sum_{i=1}^n \mathcal{A}_i^{\text{gt}} \log(\mathcal{A}_i),
\end{equation}
\end{small}
where \(\mathcal{A}_i\) represents the predicted probability distribution over tokens at position \(i\), and \(\mathcal{A}_i^{\text{gt}}\) is the ground truth token distribution.
The fine-tuned LLMs extract action instruction sets, formulated as \( A = \theta_{\text{LLM}}(c,P) = (a_1, a_2, \ldots, a_x) \), where \(P\) is the prompt used in inference, incorporating additional requirements for extracting the action instruction. 
The extracted instructions are then fed into the Act module to generate the corresponding motion.

\subsection{ACT Model}
We employ VQ-VAE~\cite{van2017neural} to discretely encode the motion, enabling efficient motion representation learning.
The encoder and decoder are denoted as \( E \) and \( D \).
For a human motion \(\mathbf{X} = [x_1, x_2, \ldots, x_t] \), the latent feature \( \mathbf{Z} = E(\mathbf{X}) \) is represented as \( \mathbf{Z} = [z_{1}, z_{2}, \ldots, z_{t/l}] \), where \( l \) is the temporal downsampling rate of the encoder \( E \).
Quantization of each latent feature \( z_i \) entails its mapping to the nearest element \( c_k \) within the codebook \( C \),defined as:
\begin{small}
\begin{equation}
\hat{z_i} = \underset{c_k \in C}{\operatorname*{\arg\min}} \|z_i - c_k\|_2.
\end{equation}
\end{small}

We train the motion VQ-VAE using the loss and training strategy similar to~\cite{t2m-gpt}.
With a learned motion VQ-VAE, a human motion sequence \(\mathbf X \) can be mapped to discrete indices \( \mathbf{I} = [i_1, i_2, \ldots, i_{t/l}, \text{End}] \),where \(i_{i}\in[1,2,\ldots,i_{t/l}]\) represents an index from the learned codebook.
Each index \( i_j \) corresponds to a codebook entry \( c_{i_j} \), defining the quantized representation for that segment of the motion sequence.
The generation is formulated as an autoregressive next-index prediction task: given previous \(t - 1\) indices (\(\mathbf{I}<t\)) and scene text condition \(c\). 
Our objective is to predict the next indices \(p(i_t|\theta_{\text{LLM}}(c),\mathbf{I}<t)\), where \(\theta\) represents the Think Model trained in the first stage, with its parameters frozen.
The training optimization goal is defined by denoting the likelihood of the full sequence as \(p(\mathbf{I}|c) = \prod_{i=1}^{T/l} p(I_i|\theta_{\text{LLM}}(c))\). 
We directly maximize the log-likelihood of the data distribution:
\begin{small}
\begin{equation}
\mathcal{L}_{trans} = \mathbb{E}_{\mathbf{I} \sim p(\mathbf{I})}[-\log p(\mathbf{I}|\theta_{\text{LLM}}(c))]. \label{formula:5}
\end{equation}
\end{small}

For the several sets of action instructions obtained, motions are generated by following an autoregressive process over each action instruction sets. For an action instruction sets \( A = \theta_{\text{LLM}}(c,P) = (a_1, a_2, \ldots, a_x) \), we define the generation of each action with the following:
\begin{small}
\begin{equation}
\mathbf{I}_{x} = 
\begin{cases} 
f\left(\{a_{x}, \text{null}\}\right) & \text{if } x = 1, \\
f\left(\{a_{x}, \mathbf{I}_{x-1}[-n:]\}\right) & \text{if } x > 1,
\end{cases}
\end{equation}
\end{small}
where \(a_x\) denotes the current action, and \(\mathbf{I}_{x-1}[-n:]\) indicates the last \(n\) indices generated by the previous action. For the first action, an empty ID list is used. For all subsequent actions, the action instruction and the most recent \(n\) indices from the prior action are used to predict the next index.
The generation for each action instruction \(a_x\) begins from the initial embedding and proceeds autoregressively until the model predicts the End token.
If an End token of the current action instruction is encountered, we replace the action instruction to update it.
Upon completion of each action segment, the indices are concatenated to form \(\mathbf{I}_{\text{total}} = [{i}_1, {i}_2, \ldots, {i}_k]\), where \(k\) is the number of action segments. Finally, the entire sequence \(\mathbf{I}_{\text{total}}\) is decoded by the VQ-VAE to generate a cohesive and smooth motion.

  

\begin{table}
	\footnotesize
	\centering
	\setlength{\tabcolsep}{3pt}
	\begin{tabular}{c|cccccc}
		\toprule
		\multicolumn{1}{c}{Task} &
        \multicolumn{1}{c}{R-Precision$\uparrow$}&
            \multicolumn{1}{c}{MM-Dist$\downarrow$}&
            \multicolumn{1}{c}{Diversity$\uparrow$} 
		\\
		\midrule
  
		Action Texts &$0.797 ^{\pm .002} $&$2.974^{\pm .008} $&$9.503^{\pm .065}$ \\
		Scene Texts & $0.665 ^{\pm .003}$  &$3.945 ^{\pm .000}$  &$8.435 ^{\pm .069}$ \\
		\bottomrule
	\end{tabular} 
      \vspace{-2mm}
      \caption{
       Comparison of metrics for GT motion in two tasks reveals poor accuracy, highlighting the failure of current metric reliability.
      }
	\label{tab:realmotion_change}
\end{table}

\begin{table*}[t] 
\centering
\setlength{\tabcolsep}{2.5pt} 
\renewcommand{\arraystretch}{0.8} 
\begin{tabular}{lcccccccccc}
\toprule
Method &Type& Train & Test & Hit Accuracy $\uparrow$ & FID $\downarrow$ & Mean Hit Distance $\downarrow$ & Diversity $\uparrow$ & MModality $\uparrow$ \\
\midrule

\multirow{2}{*}{\centering TM2T~\cite{guo2022tm2t}} & Zero-shot&\scalebox{1.5}{$\circ$} & \scalebox{1.5}{$\bullet$}& $65.5$ & $2.201 ^{\pm .020}$ & $1.407$ & $7.286 ^{\pm .075}$ & $2.600 ^{\pm .094}$ \\
&Trained & \scalebox{1.5}{$\bullet$} & \scalebox{1.5}{$\bullet$} & $68.8$ & $1.394 ^{\pm .000}$ & $1.357$ & $8.181 ^{\pm .000}$ & $2.701 ^{\pm .000}$ \\
\midrule

\multirow{2}{*}{\centering MDM~\cite{MDM}} & Zero-shot&\scalebox{1.5}{$\circ$}  & \scalebox{1.5}{$\bullet$} & $64.4$ & $0.827 ^{\pm .053}$ & $1.425$ & $8.249 ^{\pm .058}$ & $2.804 ^{\pm .052}$ \\
 &Trained& \scalebox{1.5}{$\bullet$} & \scalebox{1.5}{$\bullet$} & $73.4$ & $0.435 ^{\pm .029}$ & $1.247$ & $8.634 ^{\pm .057}$ & $2.901 ^{\pm .055}$ \\
\midrule

\multirow{2}{*}{\centering MLD~\cite{mld}} &Zero-shot& \scalebox{1.5}{$\circ$} & \scalebox{1.5}{$\bullet$} & $61.8$ & $0.897 ^{\pm .026}$ & $1.547$ & $9.289 ^{\pm .096}$ & $3.018 ^{\pm .028}$ \\
 &Trained& \scalebox{1.5}{$\bullet$} & \scalebox{1.5}{$\bullet$} & $72.4$ & $9.408 ^{\pm .060}$ & $1.313$ & $6.962 ^{\pm .063}$ & $3.086 ^{\pm .130}$ \\
\midrule

\multirow{2}{*}{\centering MotionDiffuse~\cite{zhang2022motiondiffuse}}& Zero-shot& \scalebox{1.5}{$\circ$}& \scalebox{1.5}{$\bullet$} & $70.2$ & $1.514 ^{\pm .000}$ & $1.236$ & $7.907 ^{\pm .000}$ & $1.813 ^{\pm .000}$ \\
 &Trained& \scalebox{1.5}{$\bullet$} & \scalebox{1.5}{$\bullet$}& $77.9$ & $2.688 ^{\pm .000}$ & $1.136$ & $7.703 ^{\pm .000}$ & $3.191 ^{\pm .000}$ \\
\midrule

\multirow{3}{*}{\centering T2M-GPT~\cite{t2m-gpt}} &Zero-shot& \scalebox{1.5}{$\circ$} & \scalebox{1.5}{$\bullet$}& $73.9$ & $0.516 ^{\pm .042}$ & $1.178$ & $9.396 ^{\pm .232}$ & $2.499 ^{\pm .348}$ \\
 &Trained& \scalebox{1.5}{$\bullet$} & \scalebox{1.5}{$\bullet$} & $79.6$ & $0.316 ^{\pm .015}$ & $1.081$ & $8.627 ^{\pm .080}$ & $2.620 ^{\pm .067}$ \\
\midrule

\multirow{3}{*}{\centering TAAT(Ours)} &Zero-shot& \scalebox{1.5}{$\circ$} & \scalebox{1.5}{$\bullet$} & $\textbf{75.1}$ & $\textbf{0.488}^{\pm 0.006}$
 & $\textbf{1.212}$ & $8.552 ^{\pm .095}$ & $\textbf{2.957} ^{\pm .070}$ \\
 &w/o filter& \scalebox{1.5}{$\bullet$} & \scalebox{1.5}{$\bullet$} & $78.6$ & $0.420 ^{\pm .019}$ & $1.195$ & $8.870 ^{\pm .043}$ & $3.167 ^{\pm .053}$ \\
 &Trained& \scalebox{1.5}{$\bullet$} & \scalebox{1.5}{$\bullet$}& $\textbf{79.9}$ & $0.379 ^{\pm .014}$ & $\textbf{1.075}$ & $\textbf{8.950} ^{\pm .095}$ & $\textbf{3.046} ^{\pm .070}$ \\
\bottomrule
\end{tabular}
\vspace{-2mm}
\caption{\textbf{Experiment Results on Model Zero-Shot Ability and Scene Text to Motion.} Our model demonstrates optimal performance with new scene texts in Zero-Shot experiment. After being trained on~\textsc{HumanML3D++} (Trained row), models show improved results on scene texts. Our TAAT excels in both Hit Accuracy and Mean Hit Distance, indicating a superior understanding of scene texts. Furthermore, the achieved FID and Diversity metrics suggest that we can generate high-quality motions that align well with real-world motions.}
\label{tab: combined performance}
\end{table*}

\begin{figure*}[h!]
    \centering
    \vspace{-3mm}
    \includegraphics[width=0.98\textwidth]{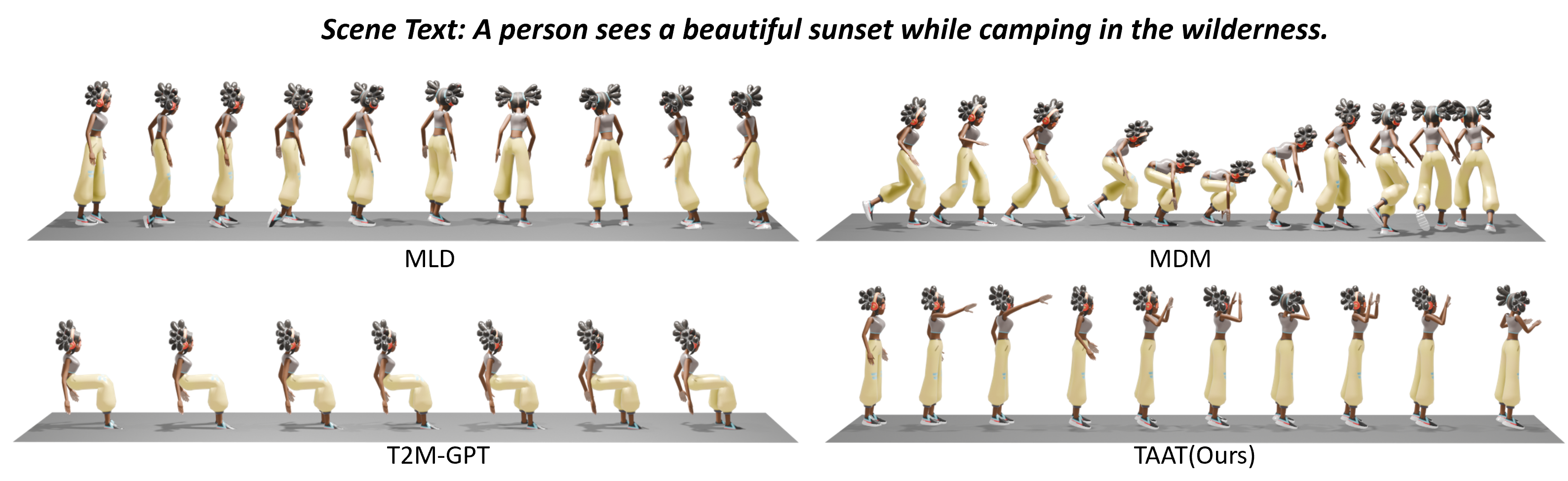}
    \vspace{-4mm}
     \caption{\textbf{Visual comparisons on scene texts.} Our model generates context-appropriate responses (pointing to the sunset, taking a photo), while other models display irrelevant actions (looking down) or remain inactive (standing still). 
 }  
    \label{fig:visual-compare_scene}
\end{figure*}

\section{Metrics}
\label{sec:metric}
Current evaluation metrics~\cite{guo2022generating} have limitations when applied to multi-solution task.
(1) R-precision and MM-Dist are limited by their requirement for the model’s output to match a single ground truth, neglecting the possibility of output motions that match the Scene Text but differ from the dataset’s GT.
(2) According to~\cite{guo2022generating}, ``R-Precision'' and ``MM-Dist'' rely on Motion \& Text feature extractors for alignment.
We retrain the feature extractors on \textsc{HumanML3D++}, with the results presented in Table~\ref{tab:realmotion_change}.
Experiments demonstrate that, when utilizing scene text as input, the existing evaluation system exhibits a misjudgment rate approaching 40\% even when the generated motions same as the ground truth, indicating a failure in the reliability of the metrics.
Inspired by QA task, we evaluate whether the model covers reference answers (HA) and measure the relevance of all generated outputs (not a single output as previous metrics) to the reference answers (MHD):

\( M \)  represents the total number of instances in the test. 
For each instance \(i\), the model generates \(N\) responses \(R_i = \{ R_i^{(1)}, R_i^{(2)}, \dots, R_i^{(N)} \} \).
\( G \) represents the ground truth, and the distance between the \( j \)-th response of instance \(R_i \) and the ground truth \( G \) is quantified by \( d_{ij} = \text{dist}(R_i^{(j)}, G_i) \).
A threshold, \( \theta \), is employed to determine whether a response qualifies as a ``hit'' based on its distance to the ground truth.
We define an indicator function \( \delta: \mathcal{R} \times \{ G \} \to \{ 0, 1 \} \) to determine whether a response meets the ``hit''criterion:
\begin{small}
\begin{equation}
\delta_{ij} = \delta(R_i^{(j)}, G_i)= 
\begin{cases} 
1, & \text{if } \text{dist}(R_i^{(j)}, G_i) \leq \theta ,\\
0, & \text{if } \text{dist}(R_i^{(j)}, G_i) > \theta .
\end{cases}
\end{equation}
\end{small}
\textbf{Hit Accuracy(HA)}: Hit Accuracy measures the ratio of hits among total data. For a given text, the model generates N responses motion, each compared to ground truth motion. A response qualifies as a hit if its distance to the ground truth exceeds a defined threshold $\theta$.
\begin{small}
\begin{equation}
\text{Hit Accuracy} = \frac{1}{M} \sum_{i=1}^M \max_{j=1,\dots,N} \delta(R_i^{(j)}, G_i).
\end{equation}
\end{small}
\textbf{Mean Hit Distance (MHD)}: Mean Hit Distance measures the average hit distance across multiple instances. For each instance, when hit, the hit distance is defined as the average distance of the qualifying responses; otherwise, a representative distance of the instance’s responses is used.
\begin{small}
\begin{equation}
\text{MHD} = \frac{\sum_{i=1}^M \sum_{j=1}^N \delta_{ij} \cdot d_{ij} + \sum_{i=1}^M \left( \mathbb{I}\left(\sum_{j=1}^N \delta_{ij} = 0 \right) \cdot \min d_{ij} \right)}{\sum_{i=1}^M \sum_{j=1}^N \delta_{ij} + \sum_{i=1}^M \mathbb{I}\left( \sum_{j=1}^N \delta_{ij} = 0 \right)}.
\end{equation}
\end{small}

We ensure a fair comparison with existing methods by employing consistent metrics for Action Texts to Motion Task.

\begin{figure*}[h!]
    \centering
    \vspace{-3mm}
    \includegraphics[width=0.98\textwidth]{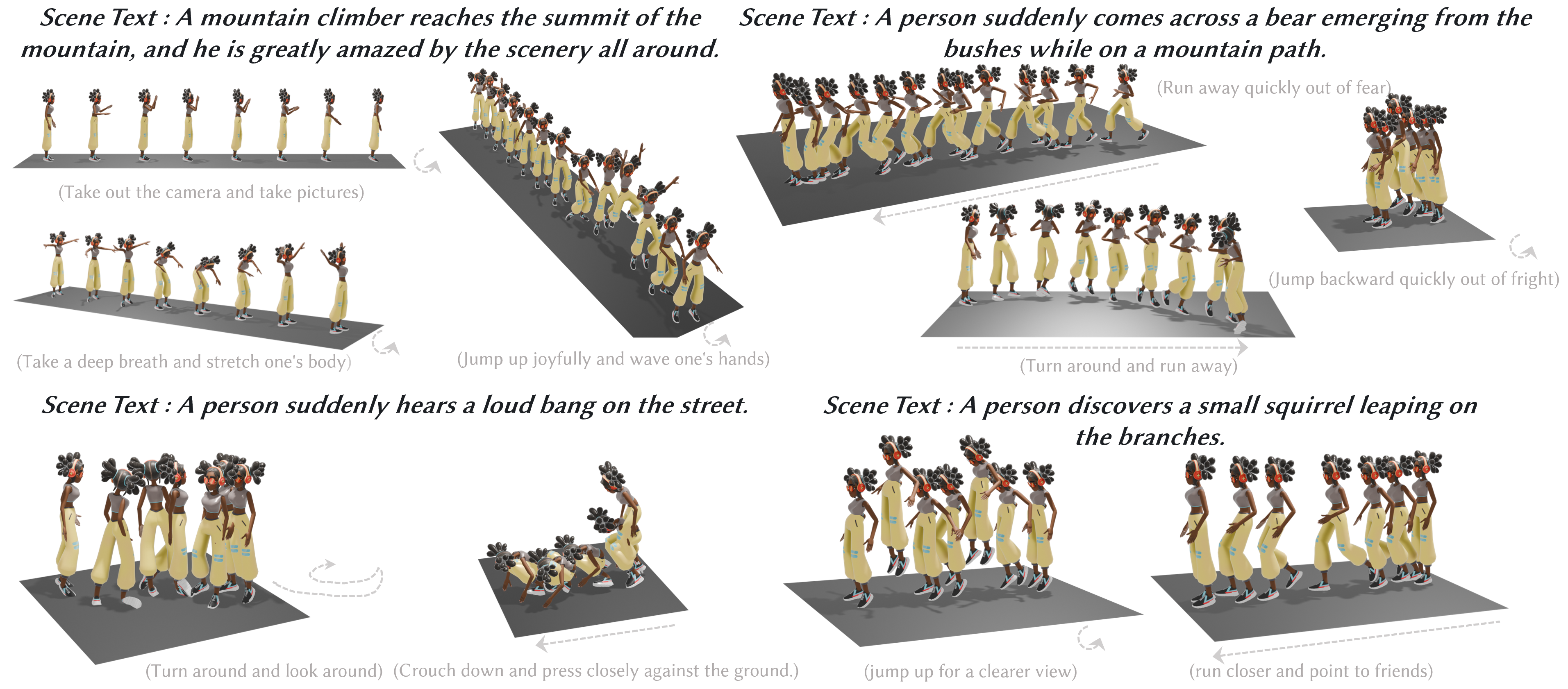}
    \vspace{-3mm}
     \caption{For a Scene Text, TAAT can generate various reasonable motions. 
 }  
    \label{fig:one_to_many_mainpaper}
\end{figure*}

\section{Experiment}

\subsection{Dataset and Implementation Details}
\textbf{Dataset.} We use the Scene Text annotations of \textsc{HumanML3D++} to benchmark the model's performance on the Scene Text to Motion, while \textsc{HumanML3D}~\cite{guo2022generating} is used to evaluate performance on the Action Text to Motion.

\noindent\textbf{Implementation Details.} 
We use the LLaMA model as the base language model and fine-tune it using LoRA. The maximum number of training epochs is set to 10, with a maximum sample size of 100,000 and a maximum input sequence length of 2048. The initial learning rate is set to 1.0e-4, and a cosine learning rate scheduler is employed, with a warmup ratio of 0.1. Additionally, the validation set is allocated 10\% of the data, and the rank for LoRA is set to 8. Given memory constraints, the batch size is set to 16. The optimization process uses the AdamW optimizer with a weight decay parameter of 0.01 to facilitate stable training.
For the generative model, we utilize 18 transformer layers with a hidden dimension of 1,024 and 16 attention heads. 
The transformer is optimized using AdamW with $\beta_1 = 0.9$ and $\beta_2 = 0.99$, a batch size of 128, an initial learning rate of $1 \times 10^{-4}$ for 150K iterations, and a decay to $5 \times 10^{-6}$ for an additional 150K iterations.

\subsection{Scene Text to Motion}
Table~\ref{tab: combined performance}, Figure~\ref{fig:visual-compare_scene} and Figure~\ref{fig:one_to_many_mainpaper} show the model's ability to learn Scene Texts and generate responsive motions. All models are retrained and tested on \textsc{HumanML3D++}.
Our TAAT demonstrates the best Hit Accuracy and Mean Hit Distance, 
indicating effective extraction of potential motions from Scene Texts and generation of contextually aligned responses.
This is achieved through scene interpretation, utilizing LLMs’ deep language comprehension and diverse candidate generation capabilities to produce multiple plausible action interpretations.
Moreover, our model achieves the highest Diversity and is ranked second in FID and MModality, indicating its ability to produce diverse realistic human poses.
T2M-GPT~\cite{t2m-gpt}exhibits poor performance in MModality, indicating that the model is still learning limited mapping relationships within the dataset. 
Consequently, it struggles to generate diverse and reasonable reactive motions for the same Scene Texts.
Note that MDM~\cite{MDM} and MotionDiffuse~\cite{zhang2022motiondiffuse} evaluate their models using the ground-truth motion length, which is impractical for real-world applications.

\subsection{Action Texts to Motion}
\begin{figure*}[ht!]
    \centering
    \includegraphics[width=0.95\textwidth]{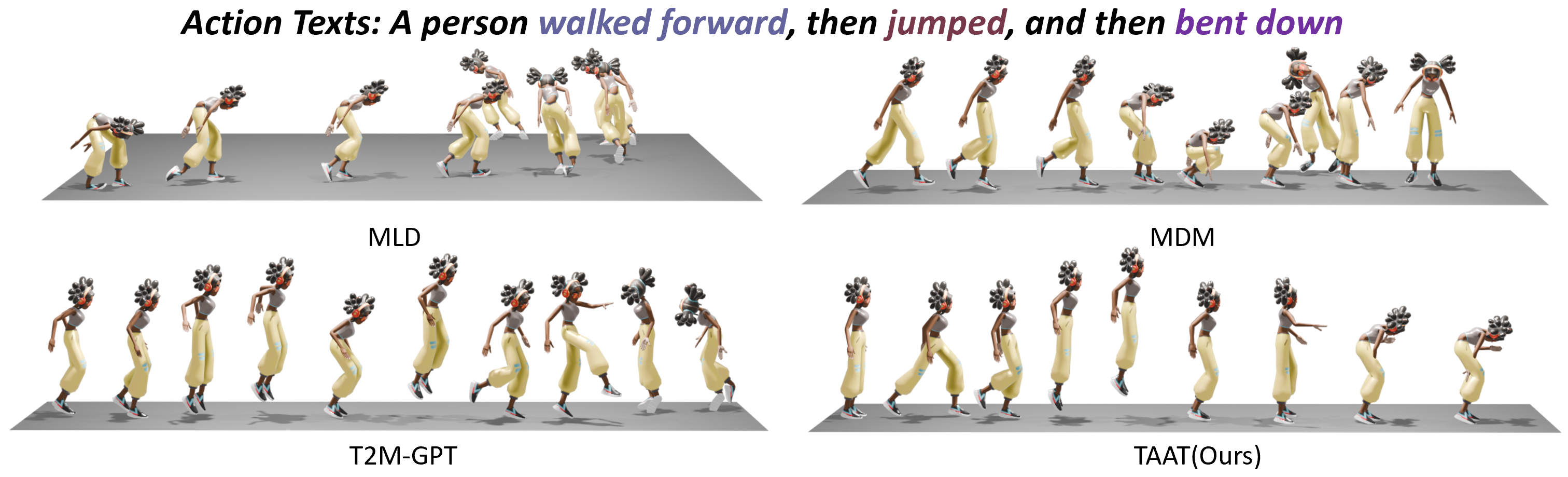}
    \vspace{-4mm}
     \caption{\textbf{Visual results on Action Texts.} Only our model performs all actions in the correct sequence, while other models exhibit issues such as missing actions (MDM~\cite{MDM}, MLD~\cite{mld}), sequence disorder (T2M-GPT~\cite{t2m-gpt}), and spatial relationship errors (MLD~\cite{mld}).
 } 
    \label{fig:visual-compare_action}
\end{figure*}

Table~\ref{tab: performance on Generation} and Figure~\ref{fig:visual-compare_action} present the results on Action Texts to Motion.
Both training and test are performed on \textsc{HumanML3D}~\cite{guo2022generating}.
In this task, we use the Think model to directly extract action instructions and generate motions. Although not specifically designed for the Action Texts to Motion, it performs well in Diversity and FID, generating diverse and realistic motions. This is achieved by independently sampling each action instruction rather than for the entire sentence, increasing the number of choices. Furthermore, as shown in Figure~\ref{fig:visual-compare_action}, the ACT model’s generation logic ensures the complete generation of all actions, suitable for multi-action generation. While diffusion-based models achieve high accuracy, they require GT motion lengths as input and cannot automatically adjust length based on text, which is impractical for real-world applications.

\subsection{Zero-shot
Experiments}
Table~\ref{tab: combined performance} illustrates our testing on the model's zero-shot capabilities.
All models are trained on the \textsc{HumanML3D}~\cite{guo2022generating}  and tested on the \textsc{HumanML3D++} to test whether the current models can understand and generate motion from Scene Texts without pretraining. 
Table~\ref{tab: combined performance} shows that existing models exhibit a decrease in multiple metrics when directly using Scene Texts inputs without pretraining.
This indicates that these models primarily learn action knowledge from the existing dataset and struggle to generalize to unseen scene texts.
Table~\ref{tab:fid-change} presents the variations in FID scores of different models across three experiments(data from table~\ref{tab: combined performance},~\ref{tab: performance on Generation}).
Despite not being trained explicitly on Scene Texts, our model achieves the best FID and Hit Accuracy and exhibits a comparatively minor decrease in FID when presented with Scene Texts inputs. 
Furthermore, it achieves the best FID and satisfactory Diversity, demonstrating the ability to generate high-quality and diverse human motions.
\begin{table}[t] 
\centering
\resizebox{\columnwidth}{!}{
\setlength{\tabcolsep}{6pt}
\renewcommand{\arraystretch}{1.2}
\begin{tabular}{lcccccc}
\toprule
Methods & R-Precision (Top-3) $\uparrow$  & FID $\downarrow$ & Diversity $\uparrow$ \\
\midrule
Real motion &$ 0.797 ^{\pm .002}$&$ 0.002^{\pm .000} $&$9.503^{\pm .065}$  \\
TM2T~\cite{guo2022tm2t}& $0.740 ^{\pm .003}$ & $1.067^{\pm .002}$ & $9.188^{\pm .002}$   \\
MDM~\cite{MDM}&$0.611 ^{\pm .007}$  & $0.544 ^{\pm .044} $ & $9.599 ^{\pm .086}$ \\
MLD~\cite{mld} & $0.772 ^{\pm .002}$ &$0.473 ^{\pm .013}$ & $9.724 ^{\pm .082}$  \\
MotionDiffuse~\cite{zhang2022motiondiffuse} & $0.782 ^{\pm .001}$& $0.630 ^{\pm .001}$  & $9.410 ^{\pm .049}$ \\
T2M-GPT~\cite{t2m-gpt}   & $0.685 ^{\pm .003}$ &$0.140 ^{\pm .006}$ & $9.844 ^{\pm 0.095}$  \\
TAAT(Ours) & $0.696 ^{\pm .003}$ &$0.461 ^{\pm .006}$ & $10.038^{\pm .095}$ \\
\bottomrule
\end{tabular}
}
\vspace{-3mm}
\caption{Experiment on Action Texts to Motion. Our TAAT works well in FID, Diversity, demonstrating that our model can generate realistic and diverse motions that are close to real human motions. Furthermore, TAAT is suitable for multi-action generation in fig~\ref{fig:visual-compare_action}.}
\label{tab: performance on Generation}
\end{table}

\begin{table}[t]
    \centering
    \footnotesize
    \setlength{\tabcolsep}{1pt}
    \renewcommand{\arraystretch}{1.2}
    \begin{tabular}{l|cccccc}
        \hline
        & TM2T~\cite{guo2022tm2t} & MDM~\cite{MDM}& MLD~\cite{mld} & MD~\cite{zhang2022motiondiffuse} & T2M-GPT~\cite{t2m-gpt} & Ours \\
        \hline
        A & $1.134$ & $0.283$ & $0.424$ & $0.884$ & $0.376$ & \textbf{$0.027$} \\
        B & $0.327$ & $-0.109$ & $8.935$ & $2.058$ & $0.176$ & \textbf{$-0.082$} \\
        \hline
    \end{tabular}
         \vspace{-2mm}
         \caption{FID variation across three experiments. Our model exhibits minimal variation in FID across different tasks. (A) shows each model’s FID change from Zero-shot to Action Text; (B) from Scene Text to Action Text.}
    \label{tab:fid-change}
\end{table}


\subsection{Ablation Study}
\textbf{Dataset Filtering.} We perform an ablation study by removing the dataset filtering, as shown by comparing the ``w/o filter'' and ``Retrain'' rows in the last two lines of Table~\ref{tab: combined performance}. Applying the filter increases Hit Accuracy and reduces Mean Hit Distance, indicating that the filtering effectively eliminates noise and irrelevant samples, thereby enhancing the overall quality and relevance of the dataset.

\noindent\textbf{Comparison of different LLMs.}
We use LLMs to construct \textsc{HumanML3D++}.
To evaluate the generation capabilities of different LLMs, we employ multiple LLMs to generate the same set of 100 scene texts based on the corresponding action texts, which are then evaluated by 23 assessors.
As performance differences are negligible, we do not factor in minor variations when selecting a LLM.
We ultimately choose Gemini considering both quality and cost.
\begin{small}
\begin{table}[h!]
\centering
\setlength{\tabcolsep}{3pt} 
\renewcommand{\arraystretch}{0.9} 
\vspace{-5mm}
\begin{tabular}{lccc}
\toprule
\cmidrule(lr){1-4}
\textbf{LLM} & \textbf{Excell.$\uparrow$} & \textbf{Sut.$\uparrow$} & \textbf{Inap.$\downarrow$} \\ 
\midrule
GPT-3.5 & 16.3\% & 74.2\% & 9.5\% \\
GPT-4o & 25.9\% & 73.7\% & 0.4\% \\
Llama-3-8B & 18.3\% & 74.6\% & 7.1\% \\
\textbf{Gemini (Ours)} & 23.5\% & 75.9\% & 0.6\% \\
\bottomrule
\end{tabular}
\vspace{-3mm}
\caption{
``Excell.,'' ``Sut.,'' and ``Inap.'' represent the ability of different models to generate scene texts from action texts, as perceived by users, corresponding to ``Excellent,'' ``Suitable,'' and ``Inappropriate'' levels, respectively.
}
\label{tab:LLM_accuracy}
\end{table}
\end{small}
\vspace{-2mm}

\noindent\textbf{Prompt design.}
Table~\ref{tab:prompt} illustrates the impact of different prompt designs on the Scene Text generation.
Acc. denotes the accuracy of the generated texts, while Score represents the human evaluation score.
Including all features (last row) achieves the highest performance, indicating that our prompt design enhances the robustness and effectiveness of the generated outputs in dataset construction. In contrast, omitting specific features, particularly Verbs or Examples, leads to a noticeable decline in output quality.

\begin{table}[h!]
    \centering
    \setlength{\tabcolsep}{2pt} 
    \renewcommand{\arraystretch}{0.9} 
    \vspace{-2mm}
    \begin{tabular}{ccccc c}
        \toprule
        w/ Verb & Quantity & Few-shot & Causality & Acc.$\uparrow$ & Score$\uparrow$ \\
        \midrule
        \scalebox{1.5}{$\circ$} & \scalebox{1.5}{$\bullet$} & \scalebox{1.5}{$\bullet$} & \scalebox{1.5}{$\bullet$} & $58.2$ &$1.80$  \\
        \scalebox{1.5}{$\bullet$} & \scalebox{1.5}{$\circ$} & \scalebox{1.5}{$\bullet$} & \scalebox{1.5}{$\bullet$} & $76.4$ & $3.75$ \\
        \scalebox{1.5}{$\bullet$} & \scalebox{1.5}{$\bullet$} & \scalebox{1.5}{$\circ$} & \scalebox{1.5}{$\bullet$} & $37.6$ & $1.40$ \\
        \scalebox{1.5}{$\bullet$} & \scalebox{1.5}{$\bullet$} & \scalebox{1.5}{$\bullet$} & \scalebox{1.5}{$\circ$} & $81.8$ & $3.80$ \\
        \scalebox{1.5}{$\bullet$} & \scalebox{1.5}{$\bullet$} & \scalebox{1.5}{$\bullet$} & \scalebox{1.5}{$\bullet$} & $98.4$ &$5.00$ \\
        \bottomrule
    \end{tabular}
         \vspace{-2mm}
        \caption{\textbf{Comparison of prompt strategies.} \textit{w/  Verb} indicates verb usage restrictions; \textit{Quantity} denotes quantity requirements; \textit{Few-shot} indicates inclusion of examples; and \textit{Causality} reflects descriptions of causal relationships.
}
\vspace{-2mm}
    \label{tab:prompt}
\end{table}

\begin{figure}[h!]
\vspace{-10pt}
    \centering
    \includegraphics[width=0.4\textwidth]{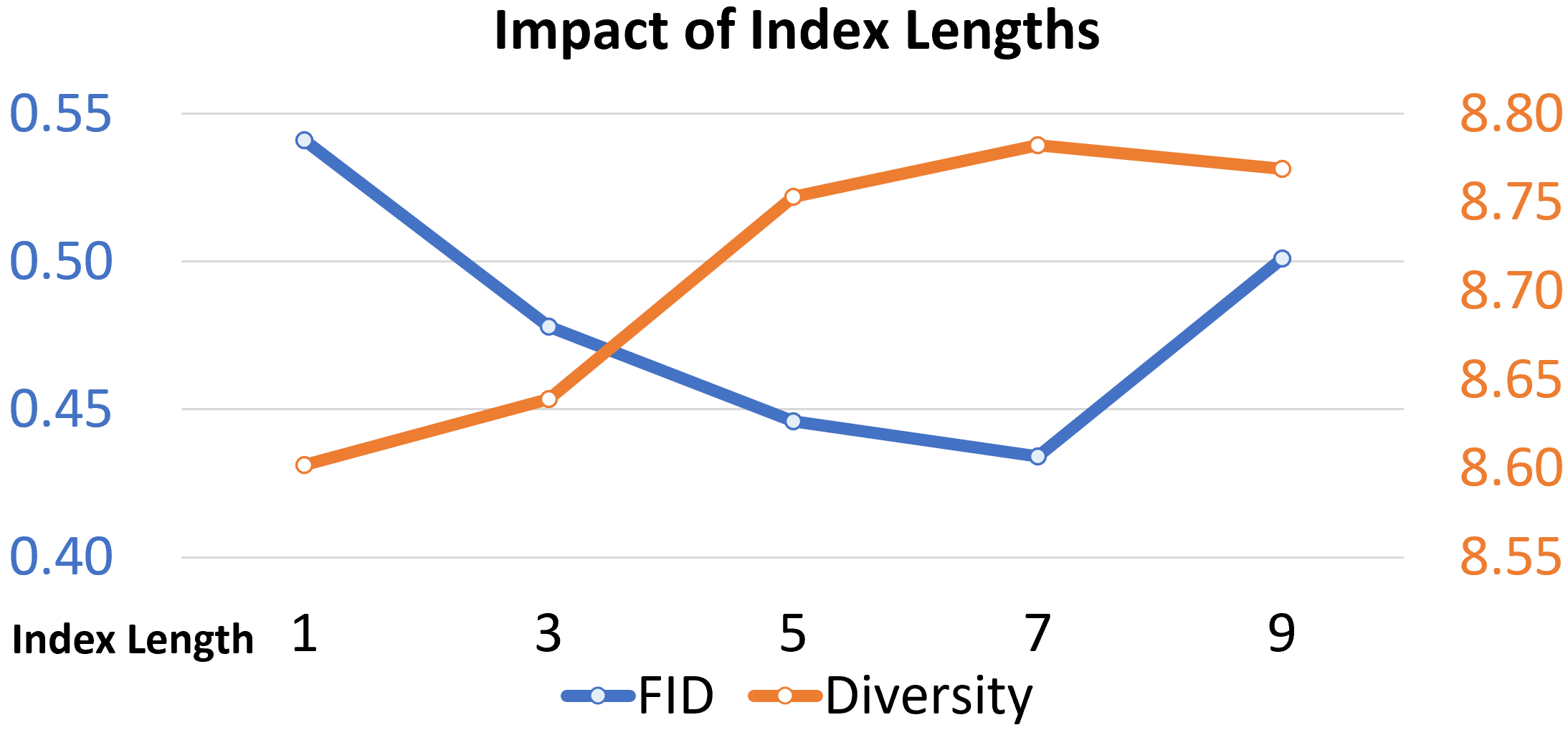}
    \vspace{-3mm}
    \caption{Impact of index lengths on motion generation.}
    \vspace{-4mm}
    \label{fig:index-change}
\end{figure}

\noindent\textbf{Index Length.}
We evaluate the impact of index lengths on the generated motions in Figure~\ref{fig:index-change} and found that an optimal index length of 7 yielded the highest generation quality. 
A shorter index often leads to discontinuities and unnatural postures, while a longer index tends to result in excessive pose repetition and increased computational costs.

\section{Conclusion}

In summary, this study introduces a novel task: inferring potential motions from Arbitrary Texts (including those with no explicit action labels), which has not been previously explored. 
Additionally, we propose a new dataset \textsc{HumanML3D++} and a more practical think-and-act framework TAAT. 
To improve evaluation, we introduce multi-solution metrics specifically designed for this novel task. 
We conduct extensive experiments to fully investigate the performance and zero-shot capabilities of existing models across the two tasks: Action Texts to Motion and Scene Texts to Motion. 
Our research establishes an essential foundation for future investigations in this domain.

{
    \small
    \bibliographystyle{ieeenat_fullname}
    \bibliography{main}
}

\end{document}